# A Study on Effect of Reference Knowledge Choice in Generating Technical Content Relevant to SAPPhIRE Model Using Large Language Model

Kausik Bhattacharya[1], Anubhab Majumder[1] and Amaresh Chakrabarti[1]

[1] Indian Institute of Science, Bengaluru, India
kausikb@iisc.ac.in

**Abstract.** Representation of systems using the SAPPhIRE model of causality can be an inspirational stimulus in design. However, creating a SAPPhIRE model of a technical or a natural system requires sourcing technical knowledge from multiple technical documents regarding how the system works. This research investigates how to generate technical content accurately relevant to the SAPPhIRE model of causality using a Large Language Model, also called LLM. This paper, which is the first part of the two-part research, presents a method for hallucination suppression using Retrieval Augmented Generating with LLM to generate technical content supported by the scientific information relevant to a SAPPhIRE construct. The result from this research shows that the selection of reference knowledge used in providing context to the LLM for generating the technical content is very important. The outcome of this research is used to build a software support tool to generate the SAPPhIRE model of a given technical system.

## 1 Introduction

In two-part research, the authors investigate whether a Large Language Model (LLM) can be used to accurately generate the technical content relevant to a SAPPhIRE representation of technical systems. However, LLM is not explicitly trained in SAPPhIRE ontology definitions, and hallucination of LLM is a major concern in generating relevant information. The first part of this two-part research, presented in this paper, demonstrates the significance of a reference knowledge choice as the contextual knowledge in Retrieval Augmented Generation (RAG), a popular method for hallucination mitigation. The second part of the research, presented in the paper titled “Development and Evaluation of a Retrieval-Augmented Generation Tool for Creating SAPPhIRE Models of Artificial Systems” by the authors, presents a new support tool to generate technical content relevant to all the constructs of SAPPhIRE model using an LLM and RAG method. The overall research plan, following the Design Research Methodology [1], is shown in Figure 1.

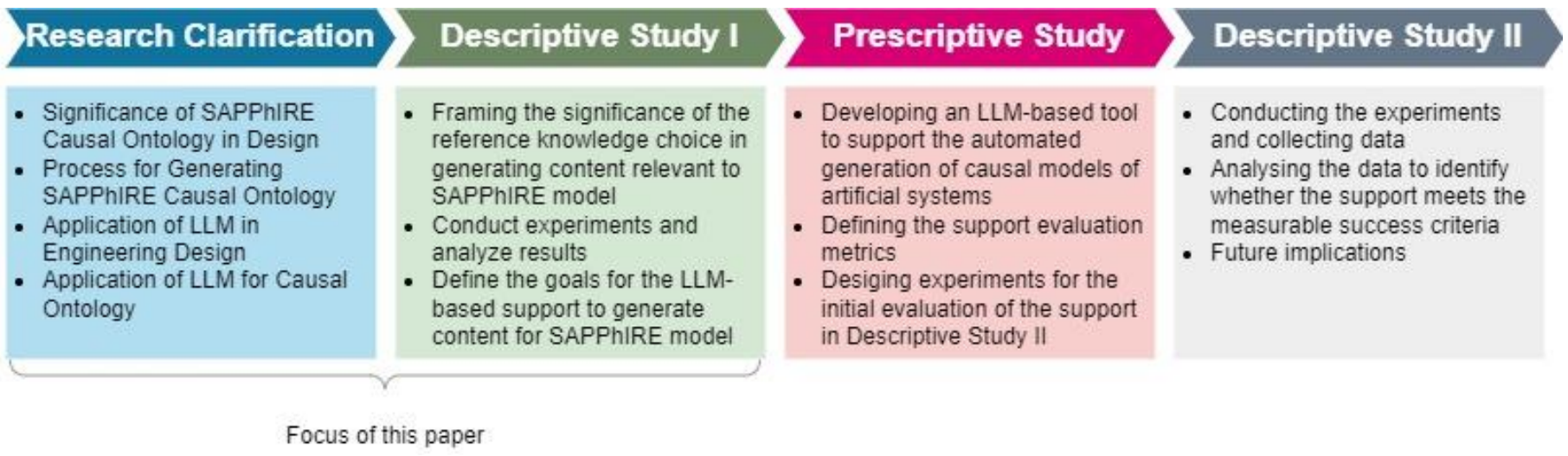

**Figure 2**. Schematic of Research Plan

# 2 Application of Large Language Models in Product Design

## 2.1 Large Language Models and Its Challenges

Large language models are at the forefront of Artificial Intelligence (AI) driven text generation, creating content that closely resembles natural human writing. Trained on extensive corpora of text, they can produce coherent and contextually relevant narratives, articles, and dialogues [2, 3]. Their advanced algorithms analyze patterns in language use, enabling them to predict and generate text with a high degree of accuracy [4]. Hallucination in large language models (LLMs) refers to the generation of incorrect or nonsensical information that the model presents as fact [5]. To mitigate these risks, it is essential to implement robust validation mechanisms and human oversight to ensure the accuracy of the model's outputs. As LLMs become more integrated into various sectors, addressing hallucinations remains a key challenge in the field of AI [6]. Research also shows that base LLMs alone are currently not sufficient for learning domain-specific ontologies [7].

## 2.2 Application of Large Language Models in Product Design

Though the application of LLM in engineering is new, its adoption is growing. Research is looking to develop useful applications with LLM in design. It is used for understanding and reasoning in Bio-inspired Design [8], in design ideation using TRIZ [9] and FBS-based design ideation using prompt engineering [10]. However, none of these works reported any hallucination-related issues and, therefore, did not look into hallucination mitigation. On the other hand, other research looked at different ways of addressing hallucinations in LLM. For example, a system engineering model using Object-Process Methodology is used with LLM for spacecraft design [11]. Retrieval Augmented Generation (RAG) is a popular and effective method employed in multiple applications where a domain-specific knowledge base is used to ground an LLM's responses [12, 13]. Causal ontologies like FBS [14] or SAPPhIRE [15] are powerful in design ideation. Hence, in this research, we look into RAG-based methods for generating information relevant to SAPPhIRE model using LLM.

### 2.3 Design Representation using SAPPhIRE Model

Representation of natural or artificial systems working using the SAPPhIRE model can act as creative stimulation during design ideation [16, 17, 18]. Hence, a method was developed to extract information about systems working from natural language technical documents and represent them using the SAPPhIRE model [19]. The SAPPhIRE model [15] has seven layers of abstraction, namely, State Changes, Actions, Parts, Phenomena, Inputs, oRgans, and Effects. A SAPPhIRRE model represents the system's physical components and their interface. It describes the physical interactions between a physical component and its surroundings involving the transfer or transformation of energy, material and information with the underlying scientific law and the condition for the physical interaction. A SAPPhIRE model can produce a rich system representation and is used in the analysis and synthesis of product design [18]. The SAPPhIRE model with an example (heat transfer from a hot body to cool surrounding air) is shown in Figure 3.

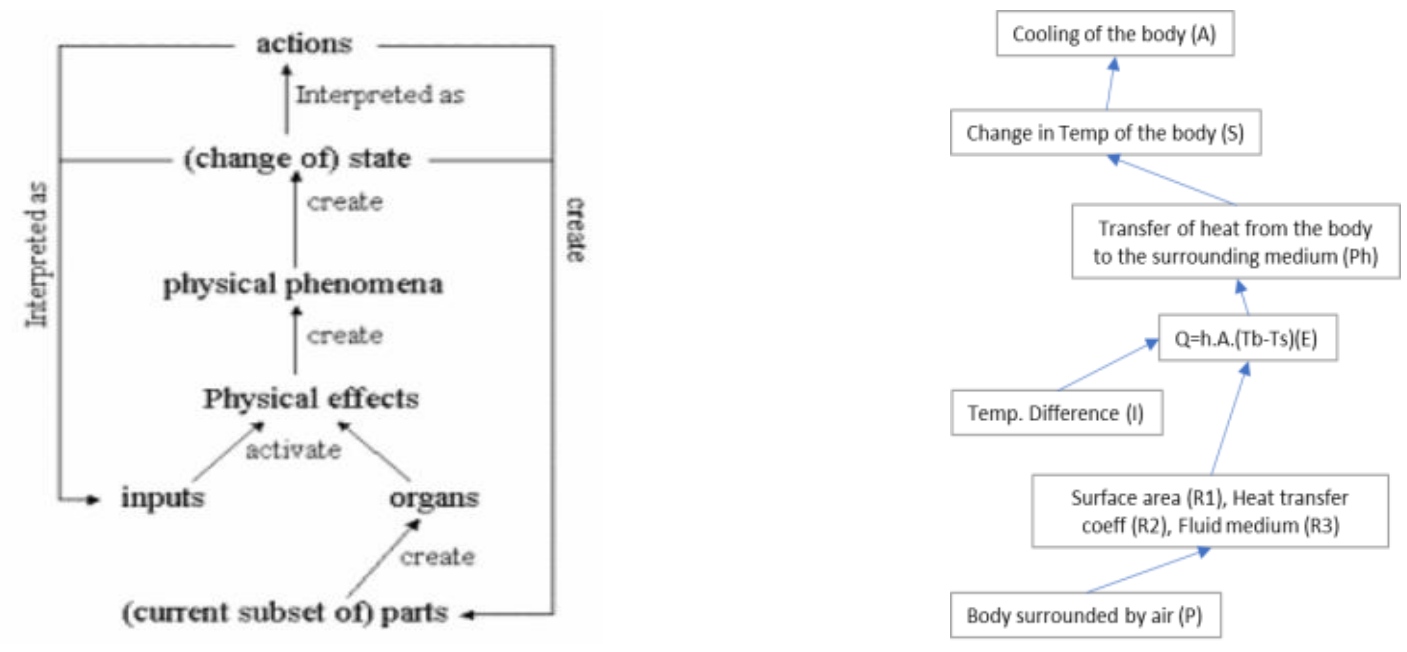

**Figure 3** – SAPPhIRE model of causality with an example of heat transfer [15]

A large language model (LLM) can be used to generate content in natural language text [2]. Hence, instead of sourcing necessary technical details from multiple documents, an LLM can be used to build SAPPhIRE model of a system. However, hallucination with large language models [5] causes undesirable inaccuracy in generated technical information. Retrieval Augmented Generation (RAG) is a method for hallucination mitigation that provides context to LLM using external knowledge. In case of content generation for SAPPhIRE model, this will produce technical information supported by scientific knowledge relevant to SAPPhIRE model constructs. In this paper, we study generating technical content relevant to the physical interaction and its condition, which corresponds to the Phenomenon and Organ constructs of SAPPhIRE model.

## 3 Research Question and Research Method

LLM's content is generated by the model, which is trained on a very large text corpus (several hundred billion tokens) with a wide variety of content from both tech-nical and non-technical domains. It is unlikely that LLM will be trained with knowledge related to the definition of the SAPPhIRE model of causality. Hence, an LLM needs appropriate technical knowledge relevant to SAPPhIRE model constructs to generate responses. Therefore, the research question for which an answer is sought is the following:

> *Does the choice of reference knowledge in Retrieval Augmented Generation matter in generating technical content relevant to the constructs of the SAPPhIRE model of causality?*

Answering the above question will show whether LLM can be used to generate technical content relevant to a SAPPhIRE construct. It is assumed that reference knowledge used as the 'context' will ground the LLM responses to scientific information relevant to a SAPPhIRE construct, and some reference knowledge will be better than others. Out of multiple 'contexts' used in RAG, one of them is assumed to be an accurate scientific expression and hence chosen as the ground truth to measure grounded-ness of LLM response. The grounded-ness of a response is calculated using the cosine similarity of two document embeddings [20], corresponding to (a) an LLM response and (b) the ground truth. The cosine similarity score henceforward will be called the similarity score for brevity. Figure 4 shows the numerical experiment plan.

Three sets of numerical experiments are conducted to find the answer to the above question. The objective of the first numerical experiment is to verify the effect of a 'context.' Hence, a one-way ANOVA test is conducted with two sets of contents specific to a given SAPPhIRE construct generated from an LLM with and without 'contexts.' The objective of the second numerical experiment is to verify if one 'context' is better than the others. The average similarity scores of the three answer groups were compared, where the responses belonging to the same answer group used a common 'context.' The third numerical experiment is conducted to study the effect of the level of details of 'contexts.' A one-way ANOVA test with two sets of LLM responses is generated with two 'contexts' of varied levels of detail is conducted. The details of the experiments are given in section 5.

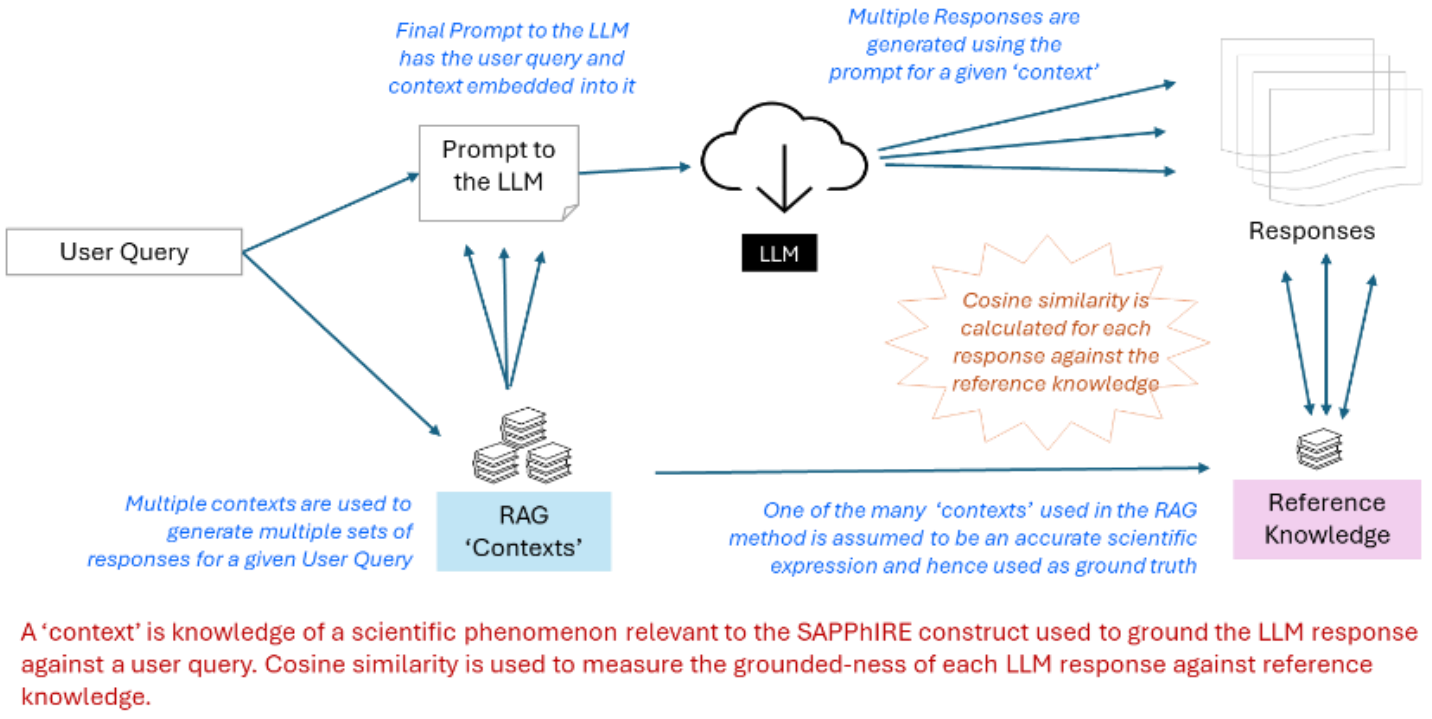

**Figure 4**. Schematic of Numerical Experiment Plan

# 4 Generating Technical Content Relevant to SAPPhIRE Model

By integrating external knowledge sources, RAG provides a ‘context’ to the LLM, ensuring that the information they produce is grounded in reference knowledge [21]. Thus, the RAG method allows LLMs to access information beyond their training data. This research implemented a computer program for content generation with the RAG method using (a) ChatGPI API [22] calls via (b) LangChain API [23] and (c) Chroma vector database API [24]. Temperature was set at 0.0 in ChatGPT API calls with the ‘GPT-4-Turbo’ model. The final prompt to the LLM includes the user query and the ‘context’ embedded into it. The user query is the ‘ask’ to the LLM to generate examples of a named scientific phenomenon and explain the physical interaction involved and its condition. Scientific knowledge regarding the physical interaction and its condition is supplied as the ‘context’ via RAG method in the LLM prompt.

# 5 Study on the Effect of ‘Context’ in LLM Responses

## 5.1 Numerical Experiments

Two scientific phenomena, namely (a) Vaporization and (b) Generation of EMF due to the Seebeck effect, are considered in both numerical experiments. Common scientific phenomena taught in high school physics and undergraduate courses are used so that the accuracy of the information can easily be verified.

In the first numerical experiment, for a given scientific phenomenon, technical contents are generated from the engineering domain as LLM response by using a prompt in two information categories, namely (a) Physical Interaction and (b) Condition for Physical Interaction. LLM response is generated with two different ‘contexts’ in each information category. In each ‘context,’ reference knowledge relevant to the information category is provided for RAG, and nine technical contents are generated for each given ‘context.’ One of the two ‘contexts’ in each information category is assumed to be ground truth, and a similarity score is calculated for each response against it. A one-way ANOVA test was done to check the effect of ‘context’ knowledge in LLM responses. Table 1.1 summarizes all the cases in the first numerical experiment.

In the second numerical experiment, for a given scientific phenomenon, technical contents are generated from the engineering domain in two information categories, namely (a) Physical Interaction and (b) Condition for Physical Interaction. Technical content is generated in these two information categories with three different ‘contexts.’ In each ‘context,’ reference knowledge relevant to the information category is provided for RAG. Since a physical interaction and its conditions can be expressed in many ways, three different ‘contexts’ are used. Three technical contents are generated for each given ‘context,’ and all these responses belong to the same group. Responses generated using a different ‘context’ belong to a different group. The knowledge of the ‘context’ is assumed to be the ground truth in that group, and a similarity score is calculated for each response against it. Table 1.2 summarizes all the cases in the second numerical experiment.

The third numerical experiment generates two sets of responses from the engineering domain for 'Physical Interaction' for each scientific phenomenon with two different 'contexts,' with the second 'context' having more technical details than the first one. In each 'context,' reference knowledge relevant to the information category is provided for RAG, and nine technical contents are generated for each given 'context. The similarity score is calculated for each response by taking the knowledge of the 'context' as the ground truth in that group. A one-way ANOVA test was done to check the effect of 'context' knowledge in LLM responses. Table 1.3 summarizes all the cases in the first numerical experiment.

**Table 1.1**. List of Test Conditions in the First Numerical Experiment

| Scientific Phenomenon | Information Category | Key Phrase Given in the Context (Three response texts are generated for each given context) |
|---|---|---|
| Vaporization | Physical Interaction | Context 1: Rapid phase transition of liquid |
| | | Context 2: None (*this can be treated as without context)* |
| | Condition for the Physical Interaction | Context 1: The liquid reaches the boiling point temperature. |
| | | Context 2: None |
| EMF Generation due to Seebeck Effect | Physical Interaction | Context 1: Generation of electric potential differences between two ends of a wire |
| | | Context 2: None |
| | Condition for the Physical Interaction | Context 1: A temperature gradient between the ends of the circuit |
| | | Context 2: None |

Note: Context 2 can be treated as 'without context'

**Table 1.2**. List of Test Conditions in the Second Numerical Experiment

| Scientific Phenomenon | Information Category | Key Phrase Given in the Context (Three response texts are generated for each given context) | Answer Group |
|---|---|---|---|
| Vaporization | Physical Interaction | Context 1: Rapid phase transition of liquid | Group 1 |
| | | Context 2: An increase in the kinetic energy of the molecules | Group 2 |
| | | Context 3: None | Group 3 |
| | Condition for the Physical Interaction | Context 1: The liquid reaches the boiling point temperature. | Group 1 |
| | | Context 2: The vapor pressure of the liquid becomes equal to the pressure exerted on the liquid by the surrounding environment | Group 2 |
| | | Context 3: None | Group 3 |
| EMF Generation due to Seebeck Effect | Physical Interaction | Context 1: Generation of electric potential differences between two ends of a wire | Group 1 |
| | | Context 2: Conversion of thermal energy into electrical energy | Group 2 |
| | | Context 3: None | Group 3 |
| | Condition | Context 1: A temperature gradient between | Group 1 |
| | for the Physical Interaction | the ends of the circuit | |
| | | Context 2: A gradient in charge carrier density between the two ends of the circuit created by the temperature difference | Group 2 |
| | | Context 3: None | Group 3 |

Note: Context 2 can be treated as 'without context'

**Table 1.3**. List of Test Conditions in the Third Numerical Experiment

| Scientific Phenomenon | Information Category | Key Phrase Given in the Context (Three response texts are generated for each given context) |
|---|---|---|
| Vaporization | Physical Interaction | Context 1: Increase in the kinetic energy of the molecules with the liquid undergoing phase change. |
| | | Context 2: Increase in the kinetic energy of the molecules with the liquid undergoing phase change and the surrounding medium are the material entities involved and with a supply of external energy. |
| EMF Generation due to Seebeck Effect | Physical Interaction | Context 1: conversion of thermal energy into electrical energy |
| | | Context 2: conversion of thermal energy into electrical energy. Two materials with different thermoelectric properties with their joints are the material entities involved and with a supply of external heat energy at one joint d to create a diffusion of charges. |

## 5.2 Results

*Numerical Experiment 1*: Table 2.1 summarizes the results from the ANOVA test. The 'context' knowledge used in Context-1 is taken as the reference knowledge to calculate the similarity score. Figure 3 shows the similarity scores of each response for both scientific phenomena.

**Table 2.1**. One-way ANOVA Test Results in Numerical Experiment 1

| Scientific Phenomenon | Information Category | Analysis Results | Test Statistics |
|---|---|---|---|
| Vaporization | Physical Interaction | Context-1 [*mean* = 0.441, *var* = 0.006]<br>Context-2 [*mean* = 0.297, *var* = 0.002] | F (1, 16) = 21.87, $p<.05$, $F_{critical} = 4.49$ |
| | Condition for Physical Interaction | Context-1 [*mean* = 0.362, *var* = 0.010]<br>Context-2 [*mean* = 0.186, *var* = 0.004] | F (1, 16) = 18.48, $p<.05$, $F_{critical} = 4.49$ |
| EMF Generation due to Seebeck Effect | Physical Interaction | Context-1 [*mean* =0.426, *var* = 0.006]<br>Context-2 [*mean* =0.25, *var* = 0.005] | F (1, 16) =23.25, $p<.05$, $F_{critical} = 4.49$ |
| | Condition for Physical Interaction | Context-1 [*mean* =0.438, *var* = 0.004]<br>Context-2 [*mean* =0.289, *var* = 0.004] | F (1, 16) = 24.41, $p<.05$, $F_{critical} = 4.49$ |

Similarity scores for the Physical interaction and its condition in 9 example cases

Vaporization | EMF Generation due to Seebeck Effect

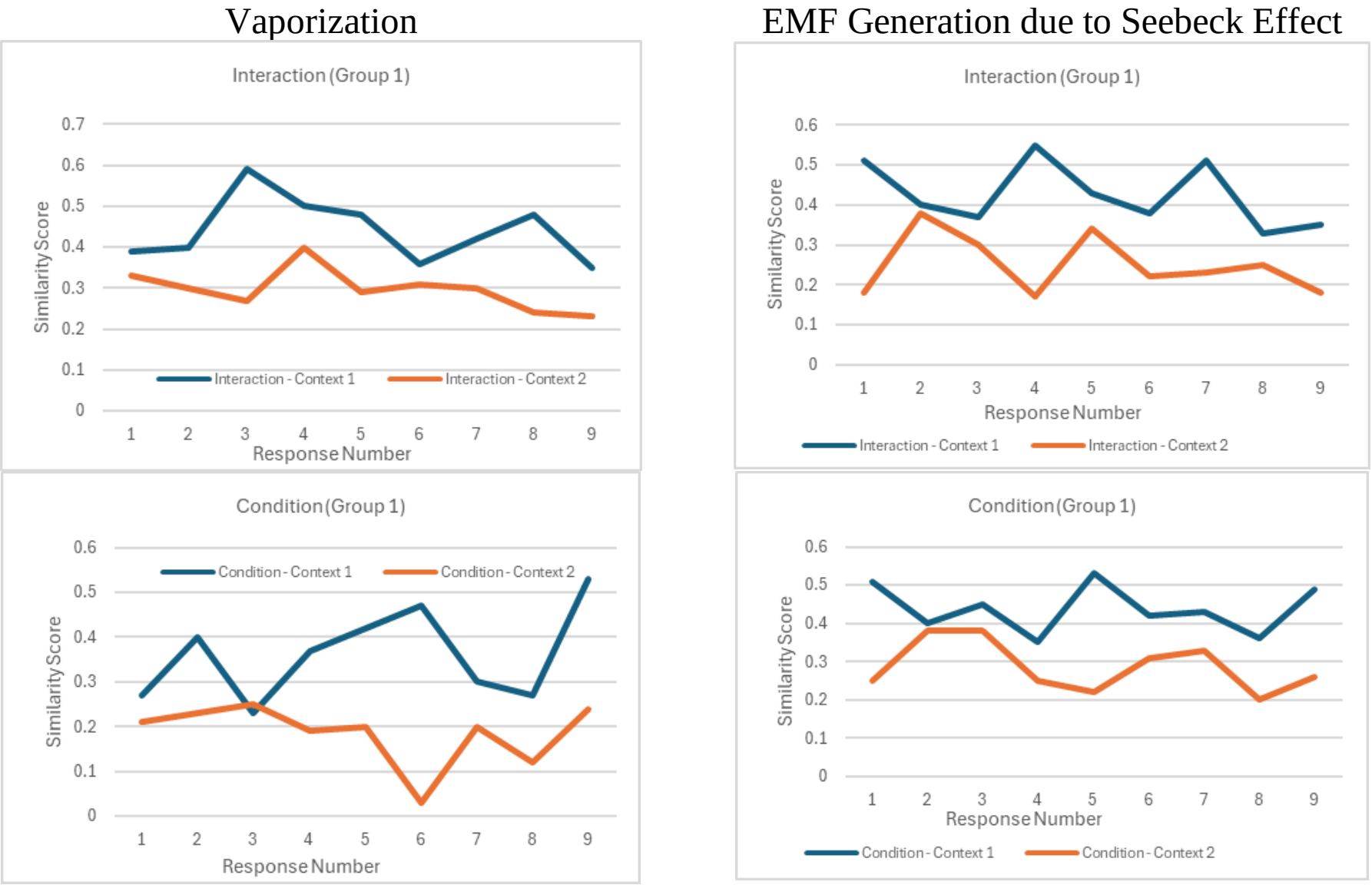

**Figure 3**. Similarity Scores of Responses in Numerical Experiment 1

*Numerical Experiment 2*: Table 2.2A summarizes the similarity scores for Vaporization and EMF Generation Due to Seebeck Effect for all the experiment cases given in Table 1.2. This experiment generates two sets of similarity scores in each information category using Context-1 and Context-2 as the ground truth. There are three groups of responses in each information category, namely, Group-1, Group-2, and Group-3. Group-1, Group-2, and Group-3 have all their responses generated using Context-1, Context-2, and Context-3 respectively. Average similarity scores are calculated for each group comprised of three responses, first with Context-1 as reference knowledge and next with Context-2 as reference knowledge. Table 2.2B shows the responses generated in each group. The key phrases in the LLM responses in each group are given in Table 2.2B. Details of the user query and LLM responses are given in the attached document available here[1].

**Table 2.2A**. Similarity scores of each answer group

| Information Category | Ground Truth for calculating the Similarity Scores | Average Similarity Scores for responses in the below groups: | | |
|---|---|---|---|---|
| | | Group-1 | Group-2 | Group-3 |
| *Scientific Phenomenon: Vaporization* | | | | |
| Physical Interaction | Context-1 | 0.45 | 0.38 | 0.27 |
| | Context-2 | 0.43 | **0.56** | 0.25 |
| Condition | Context-1 | 0.17 | 0.13 | 0.14 |
| | Context-2 | 0.18 | **0.29** | 0.18 |

[1] https://github.com/kausikbh/ICoRD25_Conf_LLM_Study_Results

| | | | | |
|---|---|---|---|---|
| *Scientific Phenomenon: EMF Generation due to Seebeck Effect* | | | | |
| Physical Interaction | Context-1 | 0.29 | 0.26 | 0.27 |
| | Context-2 | 0.23 | **0.43** | 0.29 |
| Condition | Context-1 | 0.27 | 0.30 | 0.26 |
| | Context-2 | 0.23 | **0.34** | 0.29 |

**Table 2.2B**. Responses generated in each answer group

| Response Group | Physical Interaction generated by LLM using Prompt | Condition for Physical Interaction generated by LLM using Prompt |
|---|---|---|
| *Scientific Phenomenon: Vaporization* | | |
| Group-1 | *rapid phase transition* of liquid to vapor | until it *reaches its boiling point* and vaporizes |
| | *phase transition* of the refrigerant from liquid to vapor | lowering their *boiling point and causing* vaporization |
| | *phase transition* from liquid to vapor | *reach its boiling point* and vaporize |
| Group-2 | *increase in kinetic energy* of water molecules | *vapor pressure equals* the atmospheric pressure |
| | *increase in kinetic energy* of the refrigerant molecules | *lowering their vapor pressure* to vaporize |
| | *increase in kinetic energy* of the molecules of the more volatile component | *reaches a vapor pressure* that equals the atmospheric pressure |
| *Scientific Phenomenon: EMF Generation due to Seebeck Effect* | | |
| Group-1 | *generation of an electric potential* difference | exposed to a *temperature gradient* |
| | *creation of voltage* differences | by maintaining a *temperature difference* |
| | *generation of a voltage* across | convert the *temperature gradient* between |
| Group-2 | *conversion of* thermal *energy* into electrical energy | a temperature gradient that drives a *gradient in charge carrier density* |
| | *conversion* of thermal *energy* into electrical energy | temperature difference ... *creates a charge carrier density gradient* |
| | *transformation* of thermal *energy* from the heat into electrical energy | temperature difference... *establishing the required gradient in charge carrier density* |

*Numerical Experiment 3*: Table 2.3 summarizes the results from the ANOVA test. The 'context' knowledge used in RAG of each result group is taken as the ground truth to calculate the similarity score. Figure 4 shows the similarity scores of each response group for both scientific phenomena.

**Table 2.3**. One-way ANOVA Test Results in Numerical Experiment 3

| Scientific Phenomenon | Information Category | Analysis Results | Test Statistics |
|---|---|---|---|
| Vaporization | Physical Interaction | Context-2 [*mean* = 0.421, *var* = 0.002]<br>Context-1 [*mean* = 0.353, *var* = 0.004] | $F (1, 16) = 5.56$,<br>$p = .03$,<br>$F_{critical} = 4.49$ |
| EMF Generation due to Seebeck Effect | Physical Interaction | Context-2 [*mean* = 0.237, *var* = 0.005]<br>Context-1 [*mean* = 0.147, *var* = 0.006] | $F (1, 16) =6.03$,<br>$p = 0.02$,<br>$F_{critical} = 4.49$ |

Similarity scores for the Physical interaction in 9 example cases

Vaporization EMF Generation due to Seebeck Effect

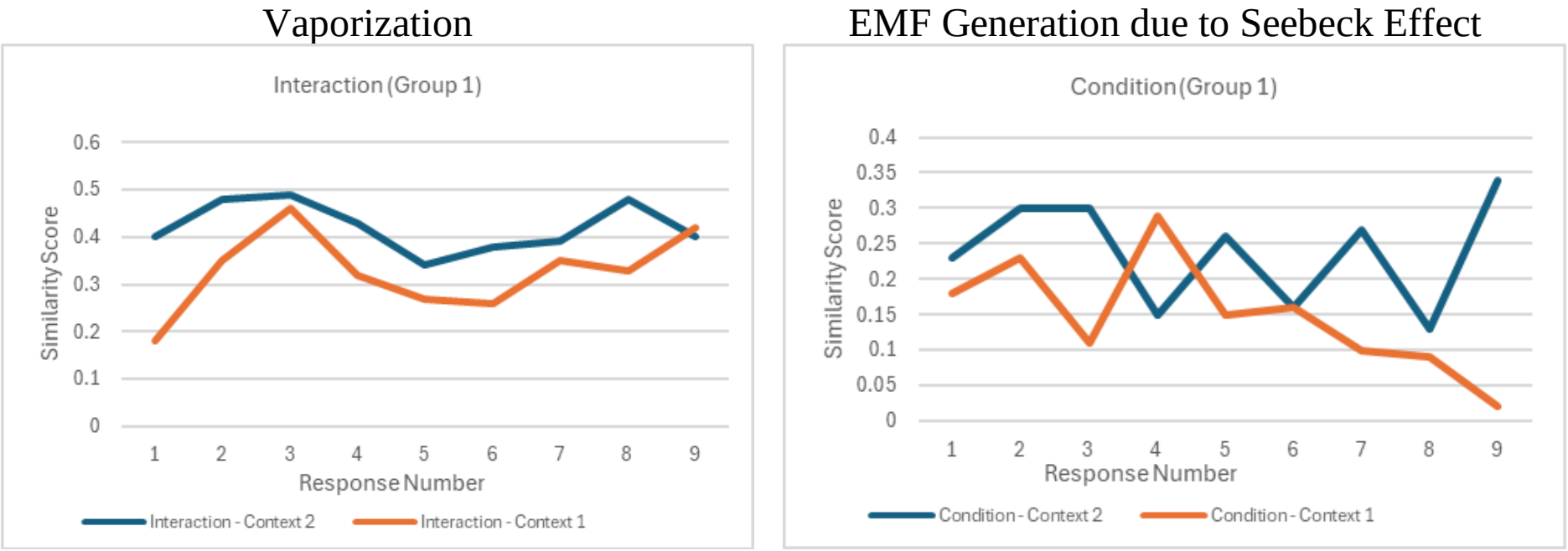

**Figure 4**. Similarity Scores of Responses in Numerical Experiment 3

## 6 Discussions and Conclusions

Since the p-value in all the test cases in Numerical experiment 1 is less than 0.05 (level of significance assumed in this experiment), we can safely reject the null hypothesis that response groups corresponding to Context-1 and Context-2 are the same. The ANOVA test results of Numerical experiment 1 show that the ‘context’ used in the RAG method influences the responses generated by the LLM. In this experiment, if Context-1 is assumed as the true knowledge, the responses generated using Context-1 in RAG will be more grounded to the truth value.

Results from Numerical experiment 2 show that LLM is generating responses based on the context provided by retrieval. The average similarity score and average similarity distance are found to be higher in Group-2. Multiple ‘contexts’ used for the same information group can be assumed to be different expressions of the scientific truth. The results show that Context-2 produces more relevant responses to the ground truth. Since scientific knowledge can be expressed in many ways, this experiment's results proved that certain descriptions of scientific knowledge will produce better responses by an LLM.

Since the p-value in all the test cases in Numerical experiment 3 is less than 0.05 (level of significance assumed in this experiment), we can safely reject the null hypothesis that response groups corresponding to Context-1 and Context-2 are the same. The ANOVA test results of Numerical experiment 3 show that the level of details in a ‘context’ used in the RAG method influences the responses generated by the LLM.

It is therefore seen in this paper that (a) with scientific knowledge provided as the context, LLM can produce rich and relevant content about the physical interaction of a system and the condition, (b) the responses generated by the LLM follow the content of the document used as ‘context,’ and (c) the choice of reference knowledge matters when generating LLM responses for SAPPhIRE construct. However, this research generated responses for only two constructs, namely, Phenomenon (i.e., physical interaction) and Organ (condition for the physical interaction), with one construct-specific short reference knowledge at a time. Therefore, in the next paper [25], we develop a support tool to generate accurate and reliable information for all seven constructs of the SAPPhIRE model using a technical document as formal reference knowledge and validate it. Since reference knowledge in LLM with RAG matters significantly in generating responses relevant to SAPPhIRE model, an accurate knowledge base of scientific laws or phenomena of natural sciences will be very useful in generating accurate ideas for design problems that conform to a causal ontology and will be looked into our future research.